\documentclass[a4paper, 10pt, conference]{ieeeconf}      
\usepackage{FG2020}
\usepackage{mathtools,amssymb}
\usepackage{footnote}
\usepackage{diagbox}
\usepackage{caption}
\usepackage{subcaption}
\usepackage{multirow}
\usepackage{tabularx}
\newcolumntype{L}{>{\raggedright\arraybackslash}X}

\FGfinalcopy 

\IEEEoverridecommandlockouts                              
\def\FGPaperID{****} 

\title{\LARGE \bf
An EEG-Based Multi-Modal Emotion Database with Both Posed and Authentic Facial Actions for Emotion Analysis
}

\author{\parbox{16cm}{\centering
    {\large Xiaotian Li$^*$, Xiang Zhang$^*$, Huiyuan Yang, Wenna Duan, Weiying Dai and Lijun Yin}\\
    {\normalsize
    Department of Computer Science, State University of New York at Binghamton, New York, USA\\
   }}
    \thanks{$^*$The first two authors contribute equally.}
}

\begin{document}

\ifFGfinal
\thispagestyle{empty}
\pagestyle{empty}
\else
\author{Anonymous FG2020 submission\\ Paper ID \FGPaperID \\}
\pagestyle{plain}
\fi
\maketitle

\begin{abstract}

Emotion is an experience associated with a particular pattern of physiological activity along with different physiological, behavioral and cognitive changes. One behavioral change is facial expression, which has been studied extensively over the past few decades. Facial behavior varies with a person's emotion according to differences in terms of culture, personality, age, context, and environment. In recent years,  physiological activities have been used to study emotional responses. A typical signal is the electroencephalogram (EEG),  which measures brain activity. Most of existing EEG-based emotion analysis has overlooked the role of facial expression changes. There exits little research on the relationship between facial behavior and brain signals due to the lack of dataset measuring both EEG and facial action signals simultaneously. To address this problem, we propose to develop a new database by collecting facial expressions, action units, and EEGs simultaneously. We recorded the EEGs and face videos of both posed facial actions and spontaneous expressions from 29 participants with different ages, genders, ethnic backgrounds. Differing from existing approaches, we designed a protocol to capture the EEG signals by evoking participants' individual action units explicitly. We also investigated the relation between the EEG signals and facial action units. As a baseline, the database has been evaluated through the experiments on both posed and spontaneous emotion recognition with images alone, EEG alone, and EEG fused with images, respectively. The database will be released to the research community to advance the state of the art for automatic emotion recognition.

\end{abstract}

\section{INTRODUCTION}
Emotions are complex responses to significant internal and external events. They are states of feeling that result in physical and psychological changes that influence our behavior \cite{Psycho11}.  They involve different components, such as subjective experience, cognitive processes, expressive behavior, psycho-physiological changes, and instrumental behavior. Most of existing research attempts to identify the emotion with one of the components, for example, behavioral signs (e.g., facial expressions) or physiological signals (e.g., EEG). Recent work (e.g., Soleymani et al \cite{Soleymani15}) has started to investigate emotion by combination of both EEG and facial expressions. However, the facial expression variation is very trivial, and there are no EEG associated with individual action units \cite{emotion_1971} included. Lack of multi-modal data with action units, expressions, and EEG signals impedes the development of the field. This motivates us to create a new database of both modalities with explicit action units and spontaneous expressions, in an attempt to explore the fusion of both external facial activity and internal brain activity in order to advance the study and understanding of emotion recognition. For context, we present an overview of the state-of-the-art databases for facial expression and physiology based emotion recognition in Table \ref{tab:database overview}.
\begin{table*}[!t]\renewcommand\arraystretch{1.2}
\caption{An overview of the state-of-the-art databases for facial expression and physiology based emotion recognition.}
\begin{center}
\begin{tabular}{|c|c|c|p{4.0cm}|p{4.0cm}|}
\hline
\multicolumn{5}{|c|}{Databases for Facial Expressions Analysis} \\
\hline
Name &  Elicitation Method   &  Subject Size      &  Modalities   &  Annotation  \\ 
\hline
CK \cite{Kanade_2000} and CK+ \cite{FER1} &  On command and Naturally occur  &  97 and 26  &  2D videos  & facial expression and action units\\ 
\hline
DISFA \cite{DISFA_2013} & Induced &  27  & 2D videos  & action units \\ 
\hline
MMI \cite{FER5}   &  On command and Induced   & 210 and 25 & 2D videos, audio, physiological signal &  facial expression and action units\\ 
\hline
BP4D+ \cite{FER4}   &  Induced    & 140  & 3D geometric facial sequences, 2D facial videos, thermal videos, physiological data sequence &  facial expression and action units\\ 
\hline
\multicolumn{5}{|c|}{Databases for Affective Analysis}  \\
\hline
SEED \cite{seed}    &  Movie induced  & 15 & 32-channel  EEG &  arousal, valence \\ 
\hline
SEED IV \cite{EmotionMeter} &  Movie induced  & 15 & 32-channel  EEG and eye track data&  arousal, valence\\ 
\hline
MAHNOB HCI \cite{hci}   &  Movie induced  & 30  &  2D videos, audio, 32-channel EEG and physiological data &  arousal, valence\\ 
\hline
DEAP \cite{DEAP_2012}   &  Music video induced & 32  &  2d facial video, 32-channel EEG and physiological data&  arousal, valence, like/dislike, dominance, and familiarity  \\ 
\hline
Ours   &  On command and induced   & 29  &  2d facial video, 128-channel EEG &  facial expression and action units, pain/relax   \\ 
\hline

\end{tabular}
\label{tab:database overview}
\end{center}
\end{table*}

{\em 1. Facial expression analysis:} Facial expression analysis (FEA) has been studied from both posed expressions and spontaneous expressions along with their facial action units.  Several existing databases have been used as testbeds for FEA (Table 1), for example, CK+ \cite{FER1}, Oulu-CASIA \cite{FER2}, MMI \cite{FER5}, BU-3DFE \cite{FER3}, BU-4DFE \cite{bu4dfe_2008}, BP4D+ \cite{FER4}, DISFA \cite{DISFA_2013}. Some of them have been used in multi-modal systems \cite{FER6} with both videos/images and other modalities (e.g., geometry, audio, 2D, 3D) for facial expressions recognition (FER), nevertheless, most of them relied on the visual information from videos or images.  For example, FER accuracy 94\%$\sim$99\% has been achieved on CK+ for 6$\sim$8 expressions \cite{FER7, FER8, FER9},  80\%$\sim$92\% with 6 classes on MMI \cite{FER12, FER11}, and  80\%$\sim$92\% on Oulu-CASIA \cite{FER12, FER9}.  It still poses a challenge in identifying emotions from facial expressions.

{\em 2. Affective analysis:} To address the issue, other modalities such as physiological signals have been employed for emotion analysis, typically, for example,  the EEG signals \cite{DEAP_2012,seed,EmotionMeter,hci}. The EEG-based emotion recognition achieved 80\% accuracy in arousal and valence \cite{emotion_1971} dimension \cite{Tripathi_2017,survey_2017}, 50\%$\sim$80\% for classifying four classes of emotions \cite{Li_2018,survey_2017}, and 64\% for classifying six emotions \cite{survey_2017}. However, it is still challenging for realizing a reliable emotion analysis from physiology data.

Multi-modal fusion from visual modality and physiology modality seems a promising method to address this issue \cite{FER14}. We propose the creation of a new database with visual data and EEG data for both posed facial actions and elicited spontaneous emotions. To verify the correlation of EEG and facial actions, we also collect EEG data associating with each action unit individually. We further validate and compare the results from single modality (e.g., expression alone, EEG alone) and multi-modality (e.g., combined expression and EEG) for recognition of both posed expressions and spontaneous emotions. 

The contribution of this work is three-fold: 
\begin{itemize}
\item This is the first data corpus with EEGs associating with individual AUs explicitly, which enables the study of relation of facial actions and EEG responses, thus allowing the information compensation as well as combination from both facial expressions and EEGs for emotion recognition.  
\item This data set contains 3 sessions including posed facial expression mimicry (6 tasks), posed action units \cite{emotion_1971} mimicry (10 tasks), and spontaneous emotion elicitation (e.g., pain and meditation). Unlike the traditional EEG-based valence-arousal analysis, we extends the EEG-based analysis for expression and action unit analysis. The link of EEG with facial actions facilitates the study of facial appearance with the brain activity, and potentially inferring the affect status. 
\item The collected EEG signals show a strong correlation with facial actions and eye blinking of both posed and spontaneous expressions. EEG features are unique for individual AUs. The EMG-like artifacts and EOG-like artifacts can be used as complementary information to benefit the expression analysis. The findings leads to the fusion of facial expressions and EEGs to improve the affection analysis.  
\end{itemize}

The following sections will elaborate the data acquisition and validation, followed by a conclusion. 

\section{DATA ACQUISITION}

\subsection{Participants}

Twenty-nine participants from the authors' institute were recruited for participating in the experiment of data collection. There are 22 males and 7 females, with ages ranging from 18 to 38 years old. Ethnic Ancestries include Asian, Mid-Eastern, White and Hispanic/Latino. The ethnic distribution is showed in Table \ref{tab:ethnic distribution}. Following the IRB-approved protocol, all participants signed an informed consent form before the start of experimentations.

\begin{figure}
\begin{center}
\includegraphics[width=0.5\textwidth]{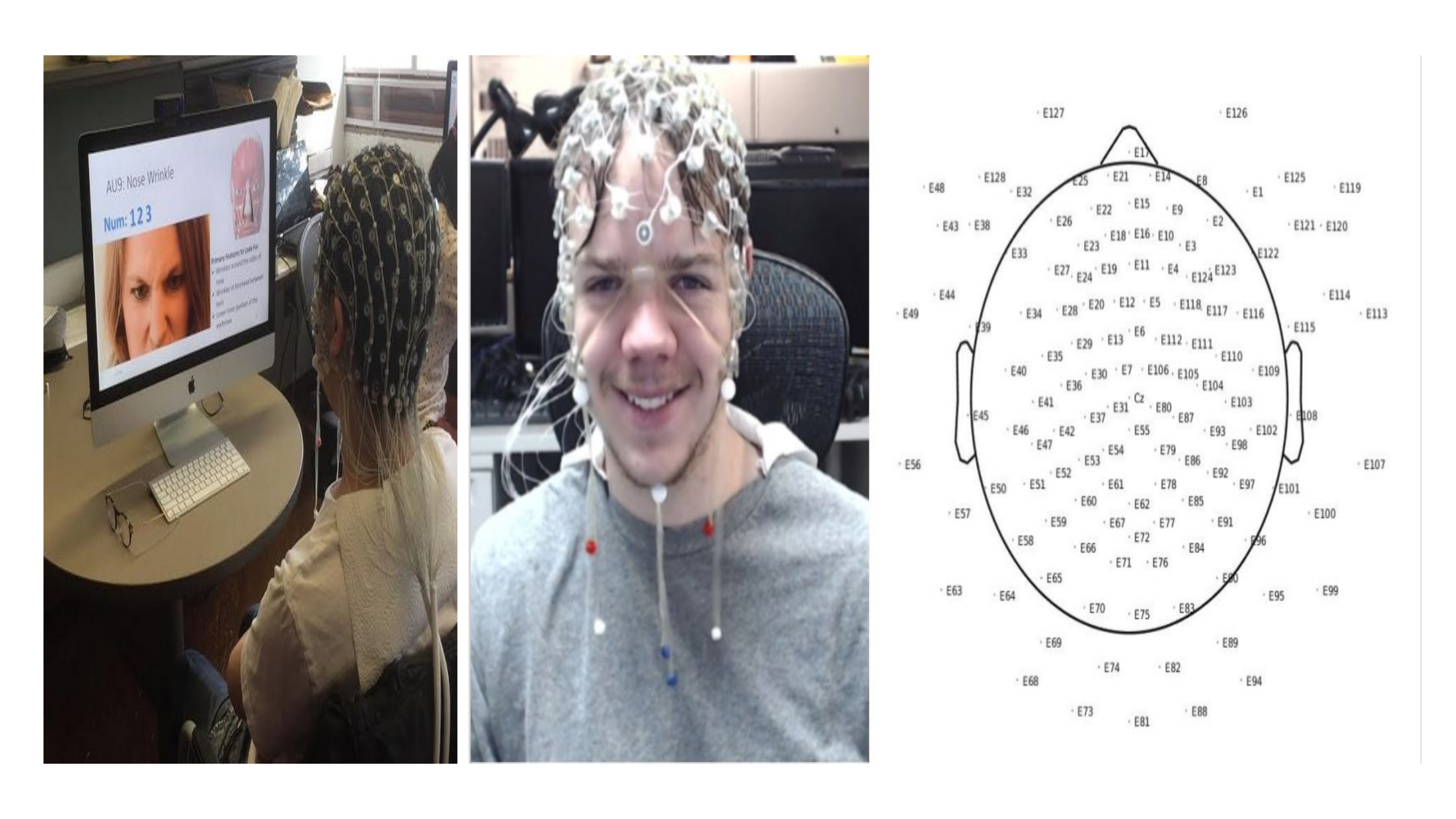}
\caption{The Experiment scene and EEG electrodes location information.}
\label{experiment_scene}
\end{center}
\end{figure}

\begin{table}\renewcommand\arraystretch{1.5}
\caption{ Ethnic distribution across 29 participants.}
\begin{center}
\begin{tabular}{|c|c|c|c|c|}
\hline
Ethnicity   & Participant Number      & Proportion   \\
\hline            
Asian &  22   &  75.9\%   \\ 
\hline
White & 2 &  6.9\%     \\ 
\hline
Mid-Eastern   &  4   & 13.8\%   \\ 
\hline
Other   &  1   & 3.45\%   \\ 
\hline

\end{tabular}
\label{tab:ethnic distribution}
\end{center}

\end{table}

\begin{figure*}[ht]
\begin{center}
\includegraphics[width=1.0\textwidth]{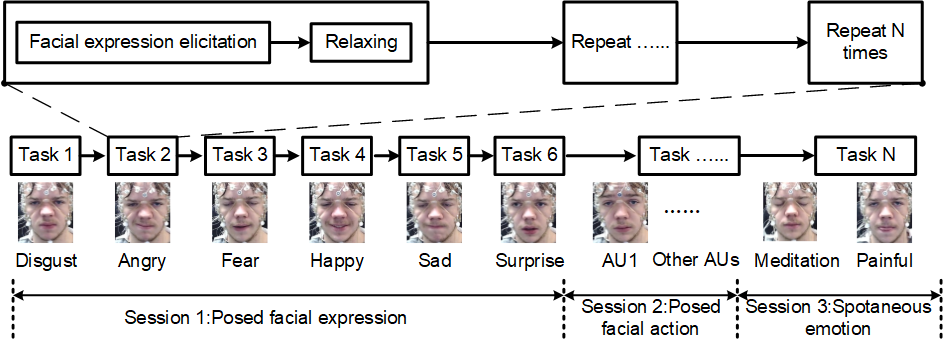}
\caption{Protocol of the data acquisition (Procedure of experiment).}
\label{EEG_aqu_process}
\end{center}
\end{figure*}

\subsection{Recording System}
The EGI's GTEN™ 100 Research Neuromodulation System \cite{GTEN_system} has been used for EEG data collection. It has 128 sensors to both record brain electrical activity and modulate it without additional sponge pads or electrodes. EGI’s Net Station Acquisition software is designed for the acquisition of dense-array EEG data.
Different from previous EEG-based affective databases, e.g., DEAP \cite{DEAP_2012} which uses International 10-20 System with 32 sensors, we used equipment with 128 sensors to record the EEG signal. The 128 sensors can cover the forehead and cheeks, providing more sensitive and reliable capture ability for facial action analysis. 
We set the sample frequency at 1000 Hz and chose the cut-off frequency for the Net Amps high-pass filter, with a 0.1 Hz cutoff by default. Fig. \ref{experiment_scene} shows the data collection at work, corresponding EEG electrodes with a frontal view, and EEG electrodes location information, respectively.

System synchronization is a critical process for data collection from multi-modality sensors. Both the EEG recording system and video recording software can generate timestamps when starting and ending the recording process. By comparing and calculating the timestamps of starting time from different sensors, we can obtain the synchronized signal accurately.

\subsection{Experiment Setup and Emotion Stimuli}
There are three sessions in the experiment for simultaneous collection of EEG signals and facial action videos, including  posed expressions, action units, and spontaneous emotions, respectively. A total of 2,320 experiment trails were recorded, which is a considerably sized database for research. The whole procedure of experiment is shown in the Fig. \ref{EEG_aqu_process}.

1) In the first session, a video illustrating six prototypical facial expressions (e.g., Anger, disgust, fear, happiness, sadness, and surprise) was shown to a participant. The participant was allowed to mimic the expressions before the start of data collection. Once the data collection starts, the participant was asked to follow the display order of six facial expressions, one-by-one, to imitate the corresponding facial expression, respectively. Each facial expression was performed three times with lasting 2$\sim$8 seconds each trial. During this time, EEG signals and facial expression videos were recorded accordingly. After an expression was imitated, and before moving on to the next expression, there were about 10$\sim$15 seconds for relax in-between.

2) In the second session, a video of displaying 10 facial action units (AU1, AU2, AU4, AU5, AU9, AU12, AU15, AU17, AU23, AU25, AU27) was shown to a participant. The participant followed the order of the 10 AUs one by one, and performed the respective facial action for 2 $\sim$ 8 seconds each. Every AU was imitated 5 times. In between two AUs, there was 5 $\sim$ 10 seconds gap to relax. 

\begin{figure}[!ht]
     \begin{subfigure}[b]{0.155\textwidth}
         \centering
         \includegraphics[width=\textwidth]{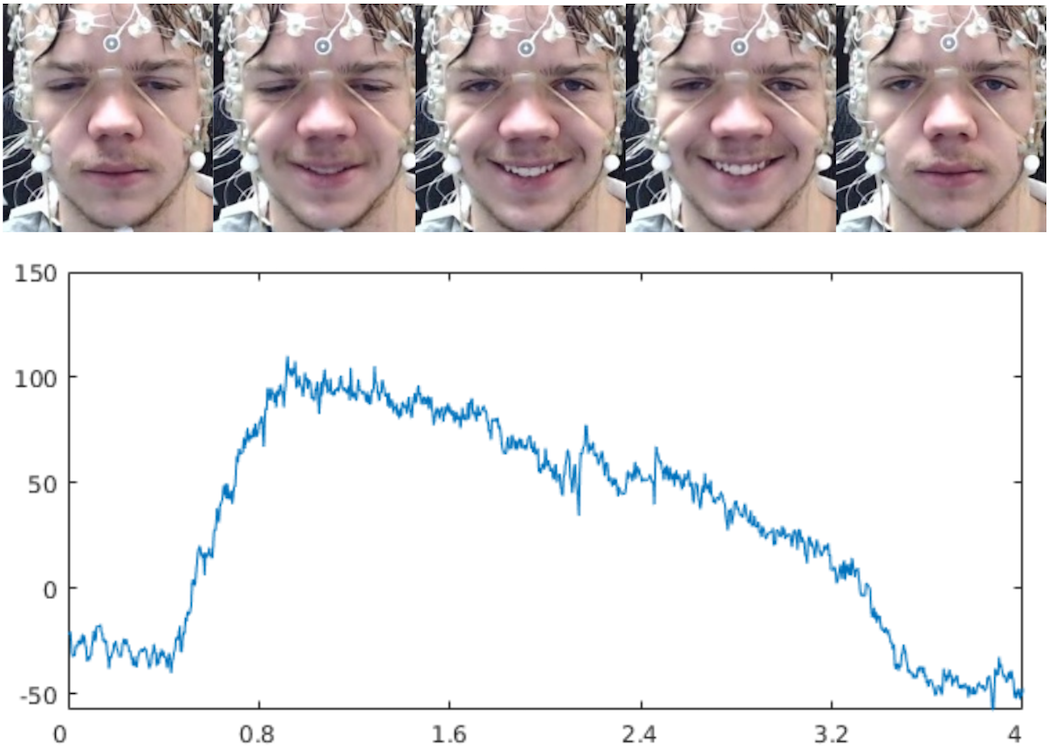}
         \caption{Happiness}
         \label{fig:sub_example_expression}
     \end{subfigure}
     \begin{subfigure}[b]{0.155\textwidth}
         \centering
         \includegraphics[width=\textwidth]{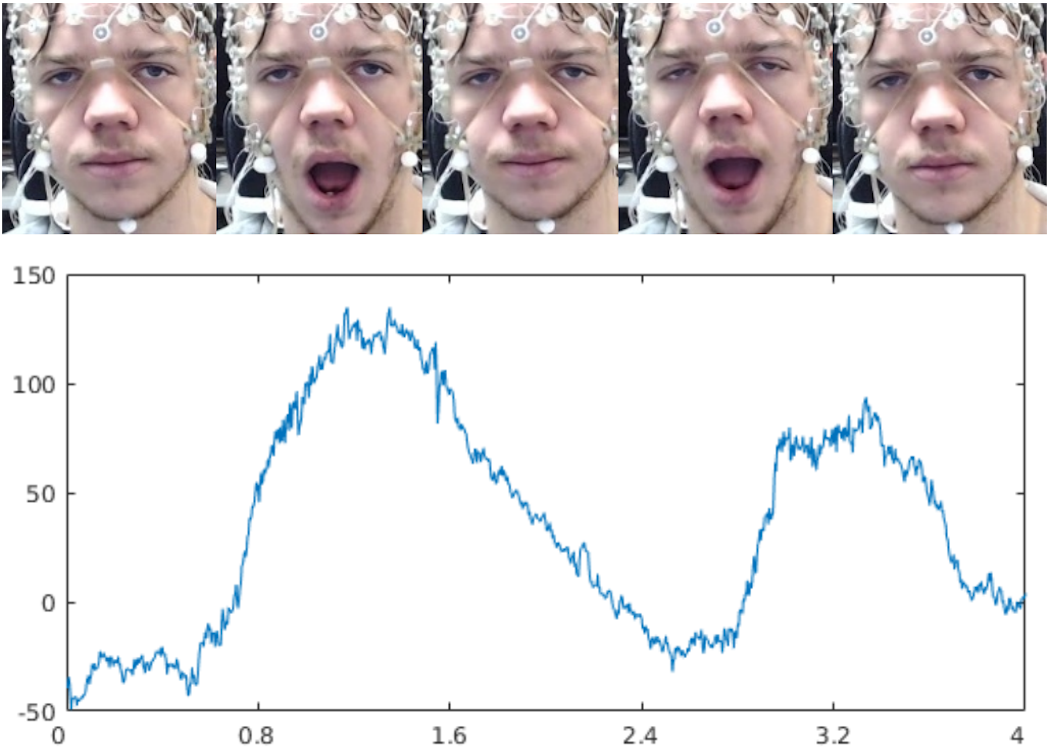}
         \caption{AU 27}
         \label{fig:sub_example_au}
     \end{subfigure}
     \begin{subfigure}[b]{0.155\textwidth}
         \centering
         \includegraphics[width=\textwidth]{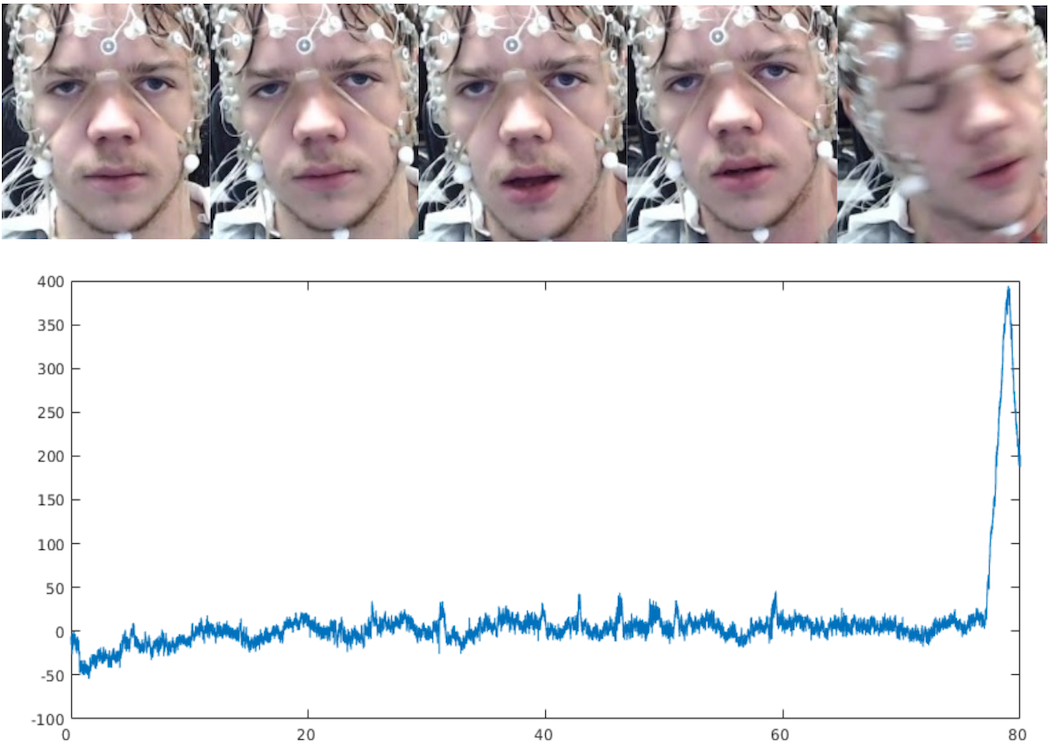}
         \caption{Pain}
         \label{fig:sub_example_pain}
     \end{subfigure}
\caption{Examples of EEG signal along with expression, action units and pain emotion. X-axis is time (seconds) sequence; Y-axis represent the mean value of the 128 location channels of EEG signals.}
\label{fig:sub_example}
\end{figure}

3) The last session collects authentic emotions elicited from two parts: meditation and pain by cold-pressor. (i) In the first part, the participant had a 5$\sim$10 seconds rest, then closed his/her eyes for meditation. During the meditation of about one minute, he/she could think of anything, and slight head or face muscle movements were allowed. (ii) In the second part, the participant was asked to submerge their hand into ice water for 90 seconds. This is to elicit an emotional response during the physical pain by the cold-pressor, followed by a self-report to rate the numerical scale of pain (0$\sim$10) by the participant. One example of a subject's signal along with face video in each session is shown in Fig. \ref{fig:sub_example}. It shows the signal significantly waved up when actions occurred, and the signal is at low level as participant felt very peaceful.

\subsection{Data Annotation and Selection}
For the posted facial expressions and action units, since the participants followed each video clip sample, their facial action and EEG signals are tagged as the designated expressions or AUs. For the spontaneous emotion (pain and neutral meditation), the self-reports are used to be the reference to rate and label the data. After the pain task, every participant was asked to provide us a brief report about the painful feeling they had, if they had one. All participants manually rate the physical pain level from 0 to 10, where 0 means no-pain and 10 means the worst possible pain. Considering the self-reports are relatively subjective judgment, we also review their corresponding video clips to annotate the data by analysing participants' facial expressions and facial actions. During the meditation task, subjects exhibit a natural state that can be viewed as a reference, which will be used as a neutral affection for the subsequent comparative study. 

Note that in order to extract the EEG signals during the event of actions or expressions, we check the video sequences and the associated EEGs sequences, and manually extract the segments of EEGs corresponding to the period of clear facial action or expressions.



\section{Data Processing and Feature Extraction}
It is necessary to pre-process the collected data before feeding the data to our baseline model. In this section, we elaborate the details for data processing and feature extraction methods.  It is worth noting that the collected EEGs' signals show a strong correlation with facial actions and eye blinking of both posed and spontaneous facial expressions. Such EMG-like artifacts (Electromyography) with frequency above 40 Hz and EOG-like artifacts  (Electrooculography) with frequency below 4 Hz reflect the facial muscle movement and eye movement/blinking \cite{EMG,emg_eog_2007,Suresh_2013}. Such a finding leads us to use all these signals for emotional facial expression analysis. Therefore, unlike the traditional EEG signal processing that removes those "noises" explicitly, it is in our belief that those peripheral signals can be used as complementary information for benefiting the analysis of both posed and spontaneous expressions.

\subsection{EEG Pre-processing and Feature Extraction}

\begin{figure}
\begin{center}
\includegraphics[width=0.5\textwidth]{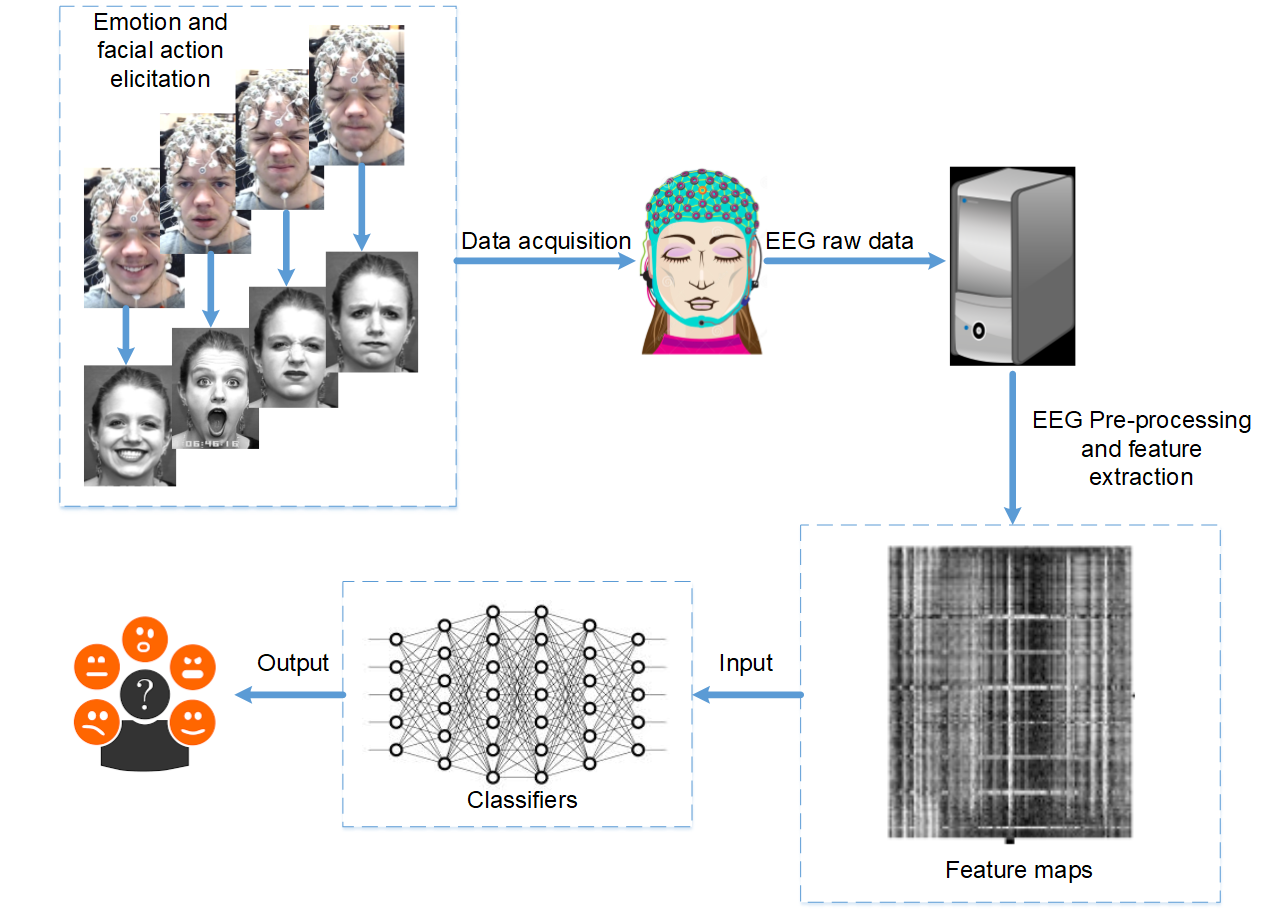}
\caption{EEG data processing, feature extraction and expression/emotion recognition.}
\label{fig:pipeline}
\end{center}
\end{figure}

In order to represent the EEG features in a two-dimensional format which is compatible to the 2D images, we take the following steps to process the EEG signals for feature extraction and feature map generation.   First, we apply a band-pass filter on the EEG data. Second, we extract the features and generate the feature map. Third, we apply the Kalman filter to smooth the extracted features map. Last, we normalize the extracted features and save them to 2D gray images as the extracted feature maps. Fig. \ref{fig:pipeline} shows a example of the generated EEG feature map. It also shows the pipeline from data acquisition to the subsequent experiment.

\subsubsection{EEG Pre-processing}

Many works \cite{weilong_2015,Bhatti_2016,Mehmood_2016} used to apply a 50 Hz band filter to extract the specific EEG frequency bands (Delta:0.1$\sim$4 Hz, Theta:4$\sim$8 Hz, Alpha:8$\sim$14 Hz, Beta:14$\sim$31 Hz, Gamma:31$\sim$50 Hz). It is to average the specific values in the corresponding frequency band respectively, which has been proved to be an efficient way for down sample. For our assumption, if EEG has the capacity to recognize the external facial expressions with the artifacts from high frequency bands, which are usually realized as noise, then the models may achieve better performance. Therefore, we apply a filter between 0.1 to 100 Hz, keeping most high frequencies information.

\subsubsection{Feature Extraction}

After applying the band-pass filter, we use the short-term Fourier transform (STFT) to extract Power spectral density (PSD) features, and calculate the Differential Entropy (DE) \cite{EmotionMeter}.

The Power spectral density function (PSD) shows the strength of the variations (energy) as a function of frequency. The unit of PSD is energy (variance) per frequency (width). The PSD features can be extracted through STFT, which is a Fourier-related transform that is used to determine the sinusoidal frequency and the phase content of the local sections of a signal as it changes over time. In this paper, we apply a 1 s time window without overlap in meditation and pain tasks. Considering these two tasks are long-time period, to balance the data for each task, we set the same time window but with 0.9s overlap in the other tasks. 

Differential entropy (DE) \cite{EmotionMeter} is used to measure the average surprisal of a continuous random variable. Its formula can be expressed as
\begin{equation}
h(X)=-\int_{X}f(x)\log(f(x))dx
\end{equation}
where X is a random variable, and $f(x)$ is the probability density function of X. In this paper, $f(x)$ is the PSD feature extracted by STFT, so we can extract DE features though,
\begin{equation}
    DE = \frac{1}{2}\log(PSD)
\end{equation}

After extracting the DE feature, we generate three different feature maps (FP), namely is Feature A, Feature B, and Feature C, respectively. Feature A is derived by the DE feature from 5 frequency bands (Delta:0.1$\sim$4 Hz, Theta:4$\sim$8 Hz, Alpha:8$\sim$14 Hz, Beta:14$\sim$31 Hz, and Gamma:31$\sim$50 Hz). Feature B is derived by 7 frequency bands (0.1$\sim$4 Hz, 4$\sim$8, 8$\sim$14 Hz, 14$\sim$31 Hz, 31$\sim$50 Hz, 50$\sim$75 Hz, and 75$\sim$100 Hz). And Feature C is same as the extracted DE feature with 100 frequency bands. Combining with 128 location channels, the three feature maps are in size of 5$\times$128, 7$\times$128, and 100$\times$128, respectively. The specification of EEG features is shown in Table \ref{tab:features}. Based on the existing works \cite{EmotionMeter,DEAP_2012,survey_2017} which shows the five bands combination (Delta, Theta, Alpha, Beta and Gamma) performs better than each individual band, we apply the whole frequency bands in our experiment, instead of testing the independent frequency band respectively. Note that Feature A is generated by the traditional method, thus it is used for comparison in the subsequent experiment.

\begin{table}
\caption{Detailed description of 3 types of Extracted EEG Features.}
\begin{center}
\begin{tabularx}{0.48\textwidth}{|L|L|L|L|} 
\hline
Feature Map Type & Feature A & Feature B & Feature C                \\ \hline
Feature Map Size & 5$\times$128 & 7$\times$128 & 100$\times$128     \\ \hline
Description & 5 frequency bands (0.1$\sim$4 Hz, 4$\sim$8 Hz, 8$\sim$14 Hz, 14$\sim$31 Hz, 31$\sim$50 Hz) & 7 frequency bands (0.1$\sim$4 Hz, 4$\sim$8, 8$\sim$14 Hz, 14$\sim$31 Hz, 31$\sim$50 Hz, 50$\sim$75 Hz, 75$\sim$100 Hz)  & 100 frequency bands (0.1$\sim$1 Hz, 1$\sim$2 Hz, 2$\sim$3 Hz...,99$\sim$100 Hz)  \\\hline
\end{tabularx}
\end{center}
\label{tab:features}
\end{table}

Afterwards, we implement the Kalman filter \cite{Kalman} for feature map smoothing. The Kalman filter has been proved to be an efficient recursive filter that estimates the internal state of a linear dynamic system (LSD) \cite{EmotionMeter} from a series of noisy measurements. 





Finally, we apply the normalization to the extracted smoothed DE feature and save it to a 2D gray image as the feature map.

\subsection{Facial Image Processing}

We record a video of every subject from a frontal camera with 24 frames per second. During the data processing, we extract all frames corresponding to the selected EEG data. We then crop the faces and resize them by 128$\times$128 using the OpenCV face detector and landmark detector.


\section{EXPERIMENTS AND VALIDATION}

\subsection{Classifiers}
In this paper, we apply three classifiers as baseline models to evaluate the performance of the new dataset: a linear SVM (EEG), a deep CNN model (static facial image), and a DANN model (both). All the experiments we conducted in this paper are with a subject-independent manner.  

\subsubsection{SVM}
We adopt the Support Vector Machine (SVM) to be the baseline algorithm for EEG based experiments. We apply it to our EEG extracted features for classification. The linear SVM kernels are imported from scikit-learn to train and evaluate the model. We search the parameter $C$ from 0.001 to 0.2 to find the optimal value.
\subsubsection{CNN}
A CNN model is deployed as a baseline model in the classification of static facial expression. We build a 4 layer convolution neural network, with each layer consisting by a conv2d, batch normalization, Relu activation, and max pooling. The out channels are set by 64, 128, 256 and 512 specifically. We set the first 3 max-pooling with kernel size 2 and the last one with kernel size 4. For the last two layers, we set a dropout for each layer. The learning rate we set up is $10^{-5}$ with learning rate decay of every 30 steps with a gamma of 0.8. We implement the CNN model through the PyTorch framework.

\subsubsection{DANN}
Considering the variant of EEG representation from different subjects, we adopt Domain-Adversarial Neural Network (DANN) \cite{DANN} as the comparison algorithm. DANN uses the idea of generative adversarial networks (GAN) \cite{GAN} which designs an adversarial process that learns both generative model and discriminative model, in which the discriminative model estimates the probabilistic distribution of a sample from a real dataset or a fake generative data set but the generative model aims to estimate the distribution of real training data. Based on the GAN method, the DANN method aims to generate domain-invariant data features that are discriminative for the classification task whereas in-discriminative for the shift between the source and target domains. In this paper, we treat the training instances as the source domain data and validation instances as the target domain data. The specification of the DANN structure is as follows: the feature extractor has 4 layers, their corresponding channels are 3 to 50, 50 to 100, 100 to 200, 200 to 400, and the kernel size is 5 $\times$ 5. We adopt max pooling to reduce the feature dimension, and their max-pooling sizes are 2, 2, 2 and 4. We also set the dropout layer to avoid overfitting, and they are located after layer 3 and layer 4. The dropout rate is 0.5. The activation function is Relu. We do batch normalization after each convolutions layer.

\subsection{Fusion strategy}
Feature level fusion (FLF) was employed to fuse the features from two modalities (facial expression images and EEG signals). Multimodal feature fusion is expected to bring more considerable performance improvement of recognition for the spacial and temporal information they carry. We concatenate the facial expression and their corresponding EEG feature map directly to form a fused feature map before feeding them into the model. Since there are three types of EEG features as shown in Table \ref{tab:features}, we  employed three types of fusion features, which are the combinations of EEG Feature A, B and C. The concatenated fusion features were resized to the same 2D dimension (128 $\times$ 128).  We used the same DANN model for a fair evaluation of the performance of both fused features and single modality features.

\begin{table*}[!ht]\renewcommand\arraystretch{1.5}
\caption{ Comparison result of single modal features and fusion features for posed expression recognition using DANN. For the specification of EEG Feature A, B and C, please refer Table \ref{tab:features}, Fusion Features A, B and C means the concatenation of facial expression with EEG Feature A, B and C. (ACC and STD means accuracy and standard deviation, respectively)}
\begin{center}
\begin{tabular}{|p{1.2cm}|p{1.2cm}|p{1.2cm}|p{1.2cm}|p{1.2cm}|p{1.2cm}|p{1.2cm}|p{1.2cm}|p{1.2cm}|}
\hline
 Evaluation method & Evaluation criteria  & EEG Feature A  & EEG Feature B   & EEG Feature C & Facial expression image & Fusion Feature A     &  Fusion Feature B   & Fusion Feature C \\\hline  
\multirow{2}{*}{LOOCV}   & ACC & 77.12\%& 80.51\%&  82.82\%   &  88.15\%   &  67.32\%   &  85.96\%   &  \textbf{95.02\%}  \\ 
\cline{2-9}   
 &STD &  0.1146  & 0.1484&  0.1255   &  0.0757   &  0.1719   &  0.0909   &  \textbf{0.0546}  \\ \hline
 \multirow{2}{*}{4 fold CV}   & ACC & 61.85\%& 66.98\%&  69.68\%   &  72.85\%   &  74.28\%   &  72.59\%   &  \textbf{76.68\%}  \\ 
\cline{2-9}   
 & STD &  0.0689  & 0.0749&  0.0819   &  0.0837   &  \textbf{0.0521}   &  0.0759   &  0.0769  \\ \hline

\end{tabular}
\label{tab:expressions_7_DANN}
\end{center}
\end{table*}

\begin{table}\renewcommand\arraystretch{2}
\caption{ The confusion matrices of fusion feature C based facial expression recognition using DANN in 4 fold cross-validation.}
\begin{center}
\resizebox{0.49\textwidth}{22mm}{
 \begin{tabular}{|c|c|c|c|c|c|c|c|}

\hline
          & Neutral  & Sadness   & Fear      & Happy & Anger     & Disgust   & Surprise \\\hline  
Neutral   & \textbf{84.51\%}& 5.94\%&  1.12\%   &  0.88\%   &  2.33\%   &  0.00\%   &  5.22\%  \\ \hline   
Sadness   &  0.71\%  & \textbf{71.71}\%&  3.00\%   &  1.12\%   &  17.88\%   &  1.12\%   &  4.47\%  \\ \hline
Fear      &  0.42\%  &  3.19\%  & \textbf{65.37}\%&  6.94\%   &  4.53\%   &  4.18\%   &  15.37\%  \\ \hline
Happy &  0.11\%  &  2.56\%   &  8.35\%   & \textbf{86.12}\%   &  2.21\%   &  0.00\%   &  0.65\%  \\ \hline
Anger     &  0.99\%  &  11.82\%   &  2.31\%   &  0.92\%   & \textbf{67.99}\%   & 8.84\%   &  7.13\%  \\ \hline
Disgust   &  0.58\%  &  4.67\%   &  4.90\%   &  6.57\%   & 15.74\%   & \textbf{64.42}\%   &  3.11\%  \\ \hline
Surprise  &  0.36\%  &  3.17\%   &  9.43\%   & 0.04\%   & 2.21\%   & 0.72\%   & \textbf{84.07}\%  \\ \hline

\end{tabular}}
\label{tab:confusion_matrix}
\end{center}
\end{table}

\begin{figure*}[ht]
\begin{center}
\includegraphics[width=1.0\textwidth]{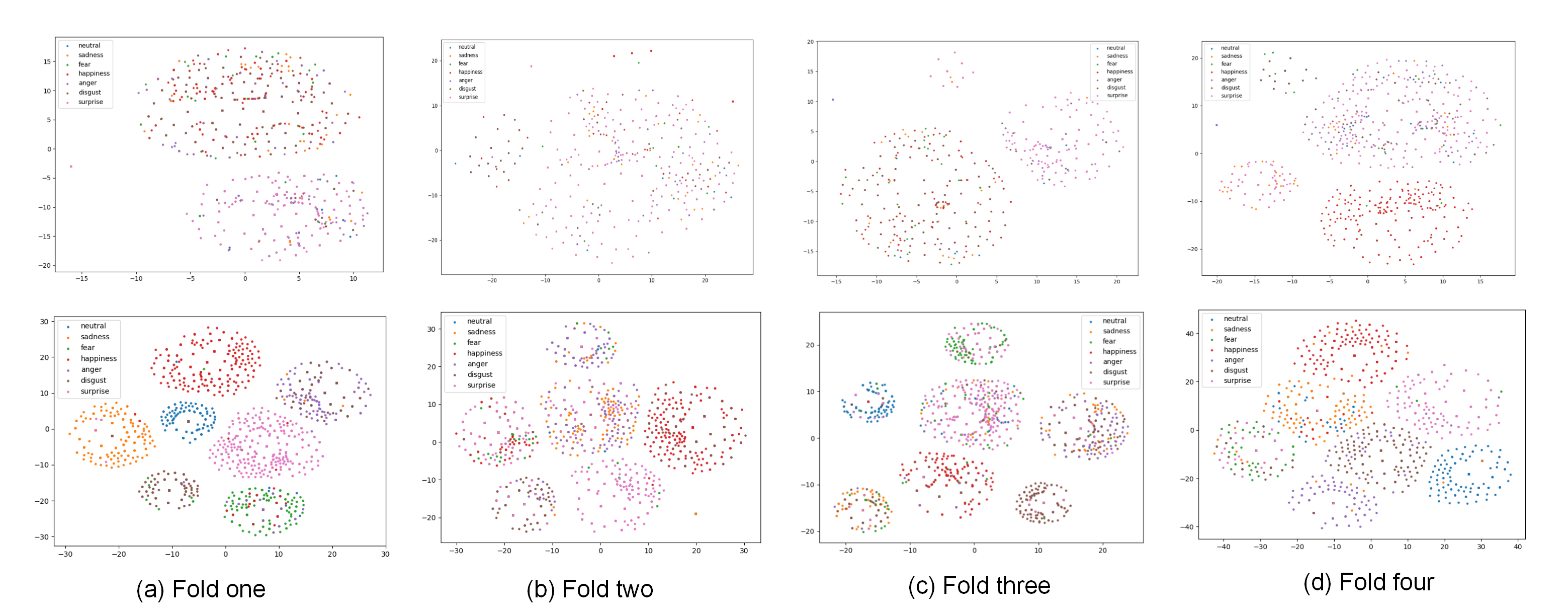}
\caption{The t-SNE visualization. The first shows the data distributions of static facial expression feature in 4 fold cross-validation from left to right, the second shows the data distribution of fusion feature trained by DANN in 4 fold cross-validation from left to right.}
\label{tSNE}
\end{center}
\end{figure*}

\subsection{Session 1: Posed facial expression recognition}

 We applied a linear SVM using three types of EEG features from Table \ref{tab:features} for 7 class facial expression recognition (Neutral, Sadness, Fear, Happiness, Anger, Disgust, and Surprise). Using leave-one-out cross-validation, the average accuracy of EEG Feature A, Feature B, and Feature C are $70.3\%$, $73.8\%$, and $71.8\%$ respectively. Accordingly, the validation results are $68.4\%$, $71.9\%$, and $71.1\%$ when the 4-fold cross-validation is applied. The CNN-based approach achieves $70.3\%$ accuracy in leave-one-out cross-validation and $68.4\%$ accuracy in the 4-fold cross-validation. The experimental results shows that DANN outperforms the linear SVM and CNN in both modalities, thus we apply the DANN to compare the single modal features and multi-modal features.


Table \ref{tab:expressions_7_DANN} shows the performance of single modal features and two-modal fused features for posed expression recognition using DANN. First, it clearly shows that EEG feature B and EEG feature C perform better than EEG Feature A because the high-frequency EEG signals (over 50HZ) are included in the feature maps. Such high-frequency EEG signals provide necessary complementary information associating with individual facial expressions to improve the classification performance significantly. Second, the performance based on the facial expression images is superior to the the performance of EEG-based single modal features by using DANN. Finally, the two-modal fusion based method generally outperforms the single modal feature based method for posed facial expression recognition. 
Fig. \ref{tSNE} shows the data distribution with 4-fold cross-validation using the t-SNE embedding method \cite{t_SNE} on facial expression images and fusion features. As we can see that the separability of fusion based method is much better than the facial expression based method. 
Table \ref{tab:confusion_matrix} illustrates the confusion matrix of the fusion method.

\subsection{Session 2:Facial action units analysis}

\begin{figure}[hbt]
    \begin{subfigure}[b]{0.50\textwidth}
        \includegraphics[width=.19\textwidth]{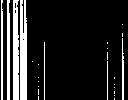}
        \includegraphics[width=.19\textwidth]{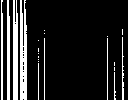}
        \includegraphics[width=.19\textwidth]{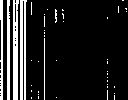}
        \includegraphics[width=.19\textwidth]{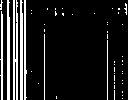}
        \includegraphics[width=.19\textwidth]{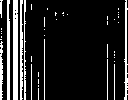}
        \\[\smallskipamount]
        \includegraphics[width=.19\textwidth]{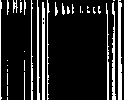}
        \includegraphics[width=.19\textwidth]{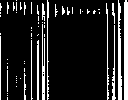}
        \includegraphics[width=.19\textwidth]{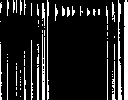}
        \includegraphics[width=.19\textwidth]{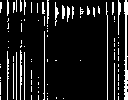}
        \includegraphics[width=.19\textwidth]{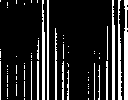}
        \caption{Feature map of average Feature C of 10 AUs respectively. X-axis is 128 sensors location, Y-axis is frequency band to 100 Hz.}
        \label{fig:au_analysis_feature}
    \end{subfigure}
    \begin{subfigure}[b]{0.5\textwidth}
        \includegraphics[width=.19\textwidth]{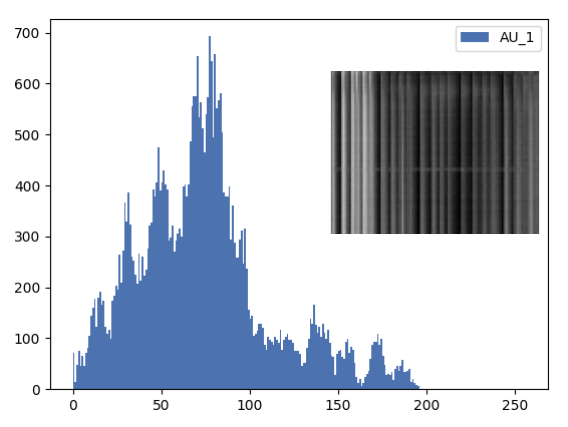}
        \includegraphics[width=.19\textwidth]{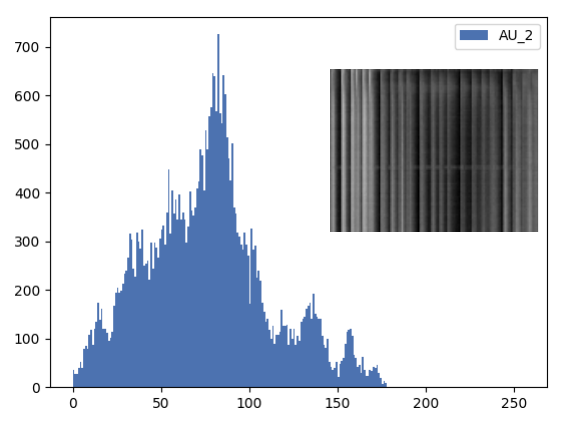}
        \includegraphics[width=.19\textwidth]{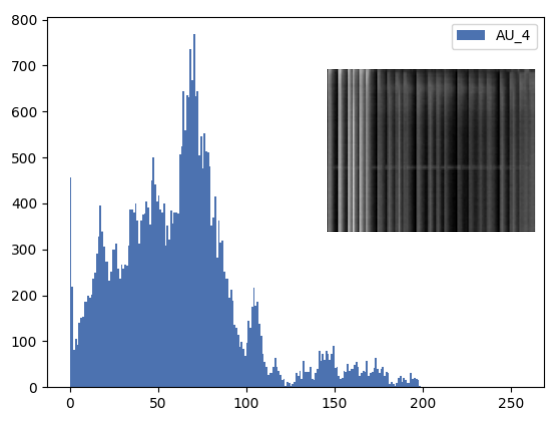}
        \includegraphics[width=.19\textwidth]{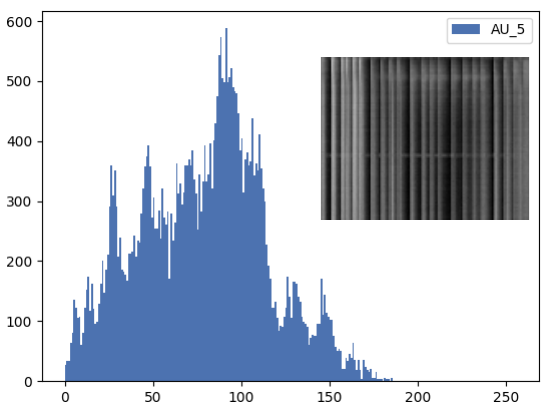}
        \includegraphics[width=.19\textwidth]{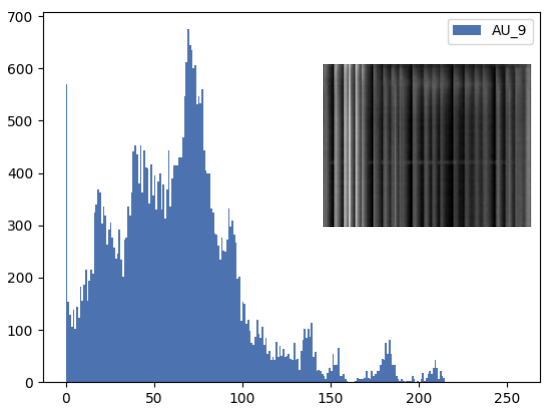}
        \includegraphics[width=.19\textwidth]{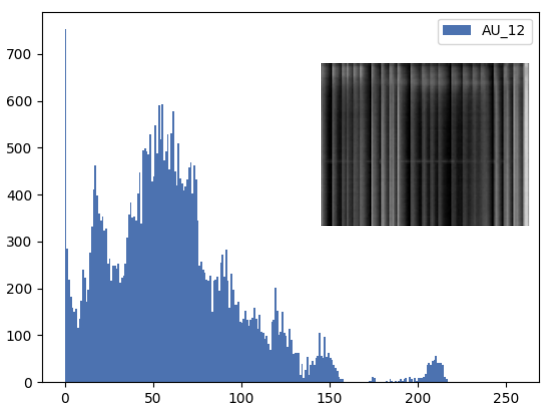}
        \includegraphics[width=.19\textwidth]{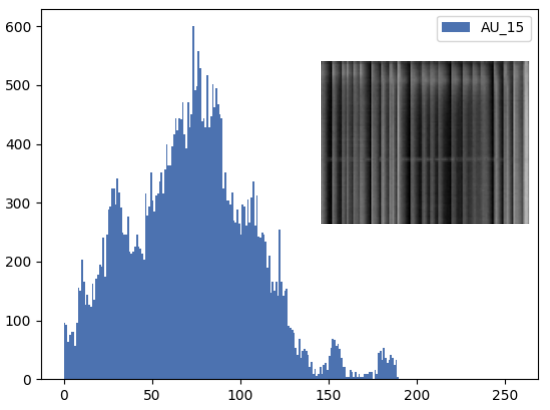}
        \includegraphics[width=.19\textwidth]{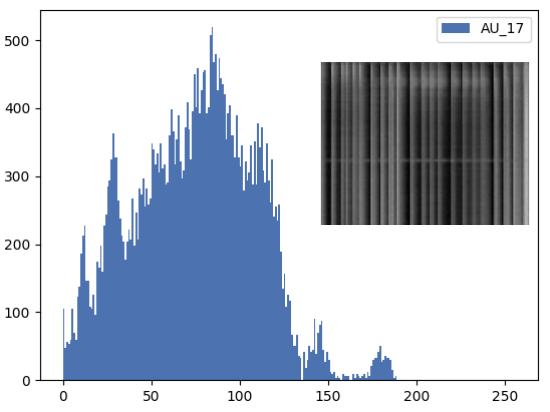}
        \includegraphics[width=.19\textwidth]{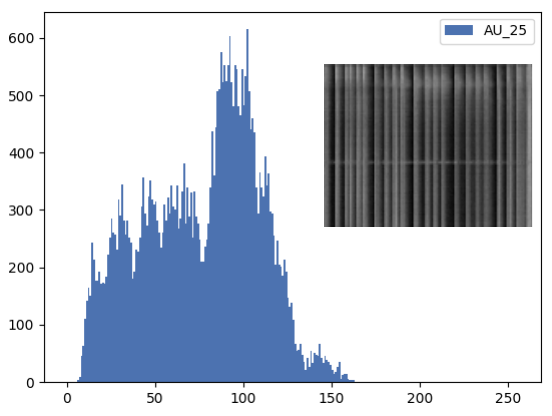}
        \includegraphics[width=.19\textwidth]{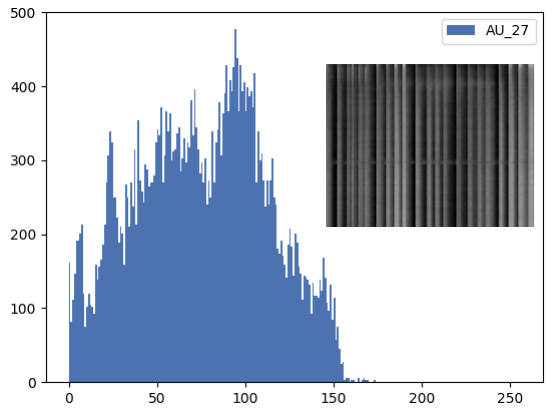}
        \caption{Histogram of average Feature C of 10 AUs respectively. Above the histogram is the corresponding Feature C.}
        \label{fig:au_analysis_histogram}
     \end{subfigure}
    \caption{ Feature map and Histogram of the Feature C from the 10 Action Units in our database. The 10 action units from left to right, row by row are: AU1, AU2, AU4, AU5, AU9, AU12, AU15, AU17, AU25, AU27.}
    \label{fig:au_analysis}
\end{figure}

To investigate the relation of AUs and the corresponding EEGs, For each AU, we calculate the average map of EEG Feature C from all subjects, and show the corresponding histogram of the corresponding average feature map.  Fig. \ref{fig:au_analysis} shows average feature map of each AU and the corresponding histogram, which are distinguishable each other. Fig. \ref{fig:au_analysis_feature} shows the average feature maps of AUs in the same part of the face have more similar patterns. For example, AU2 to AU9 are action units around eyes and nose at the upper part of a face, while AU12 to AU27 are around the mouth and chin at the lower part of face. The distribution from AU2 to AU9 are centralized around left sides in the average feature maps, which represent the lower numbers of the geodesic sensor net. AU12 to AU27 are located more in the right and middle area, meaning that the higher number of geodesic sensors are more active. Moreover, AU12 to AU25 have more activities in high frequency bands around 80Hz to 100 Hz.

In addition, in order to show the dissimilarity of any two AUs corresponding to the two feature maps, we calculate the independent (uncorrelated) coefficient of each pair of the feature maps. It is clearly shown in Table \ref{tab:action_unit}, that the 5 feature maps correspond to the 5 AUs of the upper faces have the most dissimilarity with the 5 feature maps of AUs of the lower part of the face.  For example, AU1 (Inner Brow Raiser) is most independent with AU12 (Lip Corner Puller) (e.g., 47.9\%)  but is most correlated with AU2 (Outer Brow Raiser) (e.g., 1.8\%).

\begin{table*}[htb]
\caption{ Independent (uncorrelated) coefficient matrix of the 10 Facial Action Units. The higher the more independent. }
\begin{center}
\begin{tabular}{|c|c|c|c|c|c|c|c|c|c|c|}
\hline
     & AU1 & AU2  & AU4   & AU5 & AU9  & AU12  & AU15 & AU17 & AU25 & AU27 \\\hline 
 AU1 & 0.0\%  & 1.8\%  & 7.1\%   & 5.7\% & 10.4\% & \textbf{47.9\%}  & \textbf{38.5\%} & \textbf{34.4\%} & \textbf{27.9\%} & \textbf{39.4\%}      \\\hline
 AU2 & 1.8\%  & 0.0\%  & 7.1\% & 3.4\% & 9.9\%  & \textbf{38.4\%}  & \textbf{29.5\%} & \textbf{26.0\%}  & \textbf{21.5\%} & \textbf{30.7\%}   \\\hline
 AU4 & 7.1\%  & 7.1\%  & 0.0\%    & 5.1\% & 3.0\%   & \textbf{39.5\%}  & \textbf{30.4\%} & \textbf{26.0\%}  & \textbf{18.4\%} & \textbf{33.6\%}       \\\hline
 AU5 & 5.7\%  & 3.4\%  & 5.1\%   & 0.0\%  & 7.3\%  & \textbf{29.9\%}  & \textbf{20.9\%} & \textbf{17.1\%} & \textbf{11.1\%} & \textbf{22.8\%}       \\\hline
 AU9 & 10.4\% & 9.9\%  & 3.0\%    & 7.3\% & 0.0\%   & \textbf{32.0\%}   & \textbf{24.7\%} & \textbf{21.6\%} & \textbf{16.0\%}  & \textbf{28.2\%}          \\\hline
 AU12 &\textbf{ 47.9\%} & \textbf{38.4\%} & \textbf{39.5\%} & \textbf{29.9\%} & \textbf{32.0\%} &0.0\%  &2.5\%  & 4.4\% & 12.9\% & 13.6\%         \\\hline
 AU15 & \textbf{38.5\%} & \textbf{29.5\%} & \textbf{30.4\%} & \textbf{20.9\%} & \textbf{24.7\%} & 2.5\%  & 0.0\%  & 1.2\% & 6.9\% & 8.4\%         \\\hline
 AU17 & \textbf{34.4\%} & \textbf{26.0\%}  & \textbf{26.0\%}  & \textbf{17.1\%} & \textbf{21.6\%} & 4.4\%  & 1.2\% & 0.0\%  & 4.3\% & 8.3\%       \\\hline
 AU25 & \textbf{27.9\%} & \textbf{21.5\%} & \textbf{18.4\%} & \textbf{11.1\%} & \textbf{16.0\%}  & 12.9\% & 6.9\% & 4.3\% & 0.0\%  & 9.7\%         \\\hline
 AU27 & \textbf{39.4\%} & \textbf{30.7\%} & \textbf{33.6\%} & \textbf{22.8\%} & \textbf{28.2\%}  & 13.6\% & 8.4\% & 8.3\% & 9.7\%  & 0.0         \\\hline 

\end{tabular}
\label{tab:action_unit}
\end{center}
\end{table*}

\begin{table*}[htb]\renewcommand\arraystretch{1.5}
\caption{ Evaluation of binary classes (pain versus neutral) spontaneous emotion recognition of ours database by using DANN. For the specification of EEG Feature A, B and C, please refer Table \ref{tab:features}, Fusion Features A, B and C means the concatenation of facial expression with EEG Feature A, B and C. (ACC and STD means accuracy and standard deviation, respectively.)}
\begin{center}
\begin{tabular}{|p{1.2cm}|p{1.2cm}|p{1.2cm}|p{1.2cm}|p{1.2cm}|p{1.2cm}|p{1.2cm}|p{1.2cm}|p{1.2cm}|}
\hline
 Evaluation method & Evaluation criteria  & EEG Feature A  & EEG Feature B   & EEG Feature C & Facial expression image & Fusion Feature A     &  Fusion feature B   & Fusion feature C \\\hline  
\multirow{2}{*}{LOOCV}   & ACC & 94.24\%& 95.69\%&  97.15\%   &  92.13\%   &  93.38\%   &  92.90\%   &  \textbf{98.60\%}  \\ 
\cline{2-9}   
 &STD &  0.0959  & 0.1013&  0.0860   &  0.0963   &  0.0890   &  0.1009   &  \textbf{0.0481}  \\ \hline
 \multirow{2}{*}{4 fold CV}   & ACC & 91.71\%& \textbf{94.28\%}&  90.54\%   &  86.90\%   &  87.02\%   &  92.44\%   &  91.00\%  \\ 
\cline{2-9}
 & STD &  0.0738  & 0.0443&  \textbf{0.0424}   &  0.0984   &  0.1028   &  0.0663   &  0.0637  \\ \hline
 
\end{tabular}
\label{tab:emotions_2_DANN}
\end{center}
\end{table*}

\subsection{Session 3:Spontaneous emotion recognition}
We applied a binary emotion classification for recognizing pain versus meditation.  Using leave-one-out cross-validation in linear SVM, the average accuracies of EEG Feature A, Feature B and Feature C  are $88.2\%$, $91.7\%$, and $91.9\%$, respectively. The results are $89.2\%$, $91.8\%$ and $92.3\%$ when we apply the 4 fold cross-validation. Meanwhile, through training a CNN model on the face expression images, we get accuracies of leave-one-out cross-validation and the 4 fold cross-validation with $85.2\%$ and $78.4\%$. 

Table \ref{tab:emotions_2_DANN} shows the performance of single modal features and two-modal fused features for spontaneous emotion recognition using DANN. In this part, it shows that EEG Feature B and Feature C performs generally better than EEG Feature A because the high-frequency EEG signals (over 50HZ) are included in the feature maps. We have consistent results with posed facial expression recognition showing that high-frequency EEG signals also contribute to the spontaneous emotion recognition. However, we have split results about the performance of two-modal fused features in leave-one-out cross-validation and 4 fold cross-validation (see Table \ref{tab:emotions_2_DANN}). This may caused by the conflict property of spontaneous emotion, which means participant yields some composite emotion to confuse our model. Thus, further research needs to be designed to verify our assumption in spontaneous emotion recognition.

\section{CONCLUSIONS AND FUTURE WORKS}

In this paper, we present a new EEG-based multi-modal emotion database with posed expressions, action units, and spontaneous emotions. We show the strong correlation of AUs and EEGs, and thus have applied the EEGs for AU representation. The validation result shows that the peripheral information e.g., EOG-like and EMG-like artifacts can be used as complementary features for benefiting both posed facial expression and spontaneous emotion analysis. Our validation experiments shows that the two-modality feature fusion performs better than the single-modality feature alone in terms of the facial expression classification when facial movements are not trivial. 

This work gives rise to a new investigation on how to utilize EEG signal frequency to correlate the facial behavior and emotion, with an attempt to improve the emotion analysis.  
Our future work will expand the data size to a larger scale, and will conduct EEG based AU detection and EEG-expression based fusion for AU detection. All the data will be made available to the research community.

\section{ACKNOWLEDGMENTS}

The material is based on the work supported in part by the NSF under
grant CNS-1629898 and the Center of Imaging, Acoustics, and Perception
Science (CIAPS) of the Research Foundation of
Binghamton University.


{\small
\bibliographystyle{ieee}
\bibliography{egbib}

\begin{thebibliography}{10}\itemsep=-1pt

\bibitem{survey_2017}
S.~Alarcao and M.~Fonseca.
\newblock Emotions recognition using {EEG} signals: A survey.
\newblock {\em {IEEE} Trans. on Affective Computing}, pages 374--393, 2019.

\bibitem{Bhatti_2016}
A.~Bhatti, M.~Majid, S.~M. Anwar, and B.~Khan.
\newblock Human emotion recognition and analysis in response to audio music
  using brain signals.
\newblock {\em Computers in Human Behavior}, 65:267--275, 2016.

\bibitem{FER6}
C.~Corneanu, M.~Simon, et~al.
\newblock Survey on {RGB}, 3d, thermal, and multimodal approaches for facial
  expression recognition: History, trends, and affect-related applications.
\newblock {\em {IEEE} Transactions on Pattern Analysis and Machine
  Intelligence}, 38(8):1548--1568, 2016.

\bibitem{t_SNE}
J.~Donahue, T.~Darrell, et~al.
\newblock Decaf: {A} deep convolutional activation feature for generic visual
  recognition.
\newblock {\em CoRR}, abs/1310.1531, 2013.

\bibitem{seed}
R.~Duan, B.~Lu, et~al.
\newblock Differential entropy feature for {EEG}-based emotion classification.
\newblock In {\em {IEEE} Conference on Neural Engineering}, 2013.

\bibitem{emotion_1971}
P.~Ekman and W.~Friesen.
\newblock Constants across cultures in the face and emotion.
\newblock {\em Journal of Personality and Social Psychology}, 17(2), 1971.

\bibitem{emg_eog_2007}
M.~Fatourechi, A.~Bashashati, R.~K. Ward, and G.~E. Birch.
\newblock {EMG} and {EOG} artifacts in brain computer interface systems: A
  survey.
\newblock {\em Clinical Neurophysiology}, 118(3):480--494, 2007.

\bibitem{EMG}
I.~Goncharova, D.~McFarland, T.~Vaughan, and J.~Wolpaw.
\newblock {EMG} contamination of {EEG}: spectral and topographical
  characteristics.
\newblock {\em Clinical Neurophysiology}, 114(9):1580--1593, 2003.

\bibitem{GAN}
I.~Goodfellow, J.~Pouget-Abadie, M.~Mirza, B.~Xu, D.~Warde-Farley, S.~Ozair,
  et~al.
\newblock Generative adversarial networks, 2014.

\bibitem{FER14}
H.~Gunes, B.~Schuller, et~al.
\newblock Emotion representation, analysis and synthesis in continuous space: A
  survey.
\newblock In {\em {IEEE} International Conference on Automatic Face and Gesture
  Recognition}, 2011.

\bibitem{FER7}
H.~Jung, S.~Lee, et~al.
\newblock Joint fine-tuning in deep neural networks for facial expression
  recognition.
\newblock In {\em {IEEE} ICCV}, 2015.

\bibitem{Kanade_2000}
T.~Kanade et~al.
\newblock Comprehensive database for facial expression analysis.
\newblock In {\em {IEEE} FG}, 2000.

\bibitem{DEAP_2012}
S.~Koelstra et~al.
\newblock {DEAP}: A database for emotion analysis using physiological signals.
\newblock {\em {IEEE} Trans. on Affective Computing}, 2012.

\bibitem{FER9}
C.-M. Kuo et~al.
\newblock A compact deep learning model for robust facial expression
  recognition.
\newblock In {\em {IEEE} CVPRW}, 2018.

\bibitem{Li_2018}
H.~Li et~al.
\newblock Cross-subject emotion recognition using deep adaptation networks.
\newblock In {\em Neural Information Processing}. SIP, 2018.

\bibitem{FER1}
P.~Lucey, J.~Cohn, T.~Kanade, J.~Saragih, I.~Matthews, et~al.
\newblock The extended cohn-kanade dataset ({CK+}): A complete dataset for
  action unit and emotion-specified expression.
\newblock In {\em {IEEE} CVPRW}, 2010.

\bibitem{GTEN_system}
P.~Luu, D.~Tucker, et~al.
\newblock Slow-frequency pulsed transcranial electrical stimulation for
  modulation of cortical plasticity based on reciprocity targeting with
  precision electrical head modeling.
\newblock {\em Frontiers in Human Neuroscience}, 2016.

\bibitem{DISFA_2013}
S.~M. Mavadati, J.~Cohn, et~al.
\newblock {DISFA}: A spontaneous facial action intensity database.
\newblock {\em {IEEE} Transaction on Affective Computing}, 4(2):151--160, 2013.

\bibitem{Mehmood_2016}
R.~Mehmood and H.~Lee.
\newblock A novel feature extraction method based on late positive potential
  for emotion recognition in human brain signal patterns.
\newblock {\em Computers {\&} Electrical Engineering}, 53:444--457, 2016.

\bibitem{Suresh_2013}
S.~Muthukumaraswamy.
\newblock High-frequency brain activity and muscle artifacts in {MEG}/{EEG}: a
  review and recommendations.
\newblock {\em Frontiers in Human Neuroscience}, 7, 2013.

\bibitem{FER5}
M.~Pantic et~al.
\newblock Web-based database for facial expression analysis.
\newblock In {\em {IEEE} International Conference on Multimedia and Expo},
  2005.

\bibitem{Kalman}
H.~Rauch.
\newblock Solutions to the linear smoothing problem.
\newblock {\em {IEEE} Transactions on Automatic Control}, 1963.

\bibitem{Psycho11}
D.~Schacter, D.~Gilbert, D.~Wegner, and B.~Hood.
\newblock {\em Psychology}.
\newblock Macmillan Education {UK}, 2016.

\bibitem{Soleymani15}
M.~Soleymani, S.~Asghari-Esfeden, Y.~Fu, and M.~Pantic.
\newblock Analysis of {EEG} signals and facial expressions for continuous
  emotion detection.
\newblock {\em {IEEE} Transactions on Affective Computing}, 2016.

\bibitem{hci}
M.~Soleymani, J.~Lichtenauer, et~al.
\newblock A multimodal database for affect recognition and implicit tagging.
\newblock {\em {IEEE} Transaction on Affective Computing}, 3(1):42--55, 2012.

\bibitem{FER11}
N.~Sun, Q.~Li, R.~Huan, J.~Liu, and G.~Han.
\newblock Deep spatial-temporal feature fusion for facial expression
  recognition in static images.
\newblock {\em Pattern Recognition Letters}, 119:49--61, 2019.

\bibitem{Tripathi_2017}
S.~Tripathi et~al.
\newblock Using deep and convolutional neural networks for accurate emotion
  classification on deap dataset.
\newblock In {\em AAAI}, 2017.

\bibitem{DANN}
G.~Yaroslav et~al.
\newblock Domain-adversarial training of neural networks.
\newblock {\em Advances in Computer Vision and Pattern Recognition}, 2017.

\bibitem{bu4dfe_2008}
L.~Yin, X.~Chen, Y.~Sun, T.~Worm, and M.~Reale.
\newblock A high-resolution 3d dynamic facial expression database.
\newblock In {\em {IEEE} International Conference on Automatic Face and Gesture
  Recognition}, 2008.

\bibitem{FER12}
K.~Zhang et~al.
\newblock Facial expression recognition based on deep evolutional
  spatial-temporal networks.
\newblock {\em {IEEE} Transactions on Image Processing}, 26(9):4193--4203,
  2017.

\bibitem{FER3}
X.~Zhang, L.~Yin, J.~Cohn, S.~Canavan, M.~Reale, A.~Horowitz, P.~Liu, and
  J.~Girard.
\newblock {BP}4d-spontaneous: a high-resolution spontaneous 3d dynamic facial
  expression database.
\newblock {\em Image and Vision Computing}, 32(10):692--706, 2014.

\bibitem{FER4}
Z.~Zhang, J.~Girard, Y.~Wu, X.~Zhang, P.~Liu, U.~Ciftci, S.~Canavan, M.~Reale,
  A.~Horowitz, H.~Yang, J.~Cohn, Q.~Ji, and L.~Yin.
\newblock Multimodal spontaneous emotion corpus for human behavior analysis.
\newblock In {\em {IEEE} CVPR}, 2016.

\bibitem{FER2}
G.~Zhao et~al.
\newblock Facial expression recognition from near-infrared videos.
\newblock {\em Image and Vision Computing}, 29(9):607--619, 2011.

\bibitem{FER8}
X.~Zhao, X.~Liang, et~al.
\newblock Peak-piloted deep network for facial expression recognition.
\newblock In {\em ECCV}. 2016.

\bibitem{EmotionMeter}
W.~Zheng, W.~Liu, Y.~Lu, B.-L. Lu, and A.~Cichocki.
\newblock {EmotionMeter}: A multimodal framework for recognizing human
  emotions.
\newblock {\em {IEEE} Transactions on Cybernetics}, 49(3):1110--1122, 2019.

\bibitem{weilong_2015}
W.~Zheng and B.~Lu.
\newblock Investigating critical frequency bands and channels for {EEG}-based
  emotion recognition with deep neural networks.
\newblock {\em {IEEE} Transactions on AMD}, 7(3):162--175, 2015.

\end{thebibliography}
}

\end{document}